\documentclass[sigconf]{acmart}

\usepackage{adjustbox}
\usepackage{mathtools} 
\usepackage{amsfonts}
\usepackage{bbm}
\usepackage{bm}
\usepackage{dsfont}

\usepackage[dvipsnames]{xcolor}
\usepackage{hyperref}

\usepackage{graphicx}
\usepackage[caption=false,font=footnotesize,labelfont=sf,textfont=sf]{subfig}
\usepackage{array}
\usepackage{booktabs}
\usepackage{multirow}
\usepackage{threeparttable}
\usepackage{siunitx}
\usepackage{stfloats}

\usepackage{algorithm}
\usepackage[noend]{algpseudocode}
\usepackage{tabularx}

\usepackage{listings}
\usepackage{tikz}
\usepackage{natbib}

\newcommand*\circledgreen[1]{\tikz[baseline=(char.base)]{
    \node[shape=circle,fill=ForestGreen,inner sep=0.3pt] (char) {\textcolor{white}{#1}};}}

\newcommand{\rectgreen}[1]{%
  \tikz[baseline=(char.base)]{
    \node[shape=rectangle,rounded corners=2pt,fill=ForestGreen,inner sep=1.5pt,minimum height=1.2em] (char) {\textcolor{white}{#1}};
  }%
}

\lstdefinestyle{cardcode}{
  language=Python,
  basicstyle=\linespread{0.80}\ttfamily\bfseries\footnotesize,
  keywordstyle=\color{blue!99!black}\bfseries,
  commentstyle=\color{gray!99},
  showstringspaces=false,
  breaklines=true,
  breakatwhitespace=true,
  tabsize=2,
  numbers=none,
  frame=none,
  backgroundcolor=\color{white},
  aboveskip=0.1em, belowskip=0.1em,
  xleftmargin=0em, xrightmargin=0em
}
\AtBeginDocument{%
  }

\setcopyright{acmlicensed}
\copyrightyear{2018}
\acmYear{2018}
\acmDOI{XXXXXXX.XXXXXXX}
\acmConference[Conference acronym 'XX]{Make sure to enter the correct
  conference title from your rights confirmation email}{June 03--05,
  2018}{Woodstock, NY}
\acmISBN{978-1-4503-XXXX-X/2018/06}

\begin{document}

\title{Complexity Agnostic Recursive Decomposition of Thoughts}

\author{Kaleem Ullah Qasim}
\email{kaleem@my.swjtu.edu.cn}
\orcid{0000-0002-0102-3816}
\affiliation{%
  \institution{Southwest Jiaotong University}
  \city{Chengdu}
  \state{Sichuan}
  \country{China}
}

\author{Jiashu Zhang}
\email{jszhang@home.swjtu.edu.cn}
\authornote{Corresponding Author}
\affiliation{%
  \institution{Southwest Jiaotong University}
  \city{Chengdu}
  \state{Sichuan}
  \country{China}
}

\author{Hafiz Saif Ur Rehman}
\email{saif_@msn.com}
\orcid{0009-0005-7972-729X}
\affiliation{%
  \institution{Southwestern University of Finance and Economics}
  \city{Chengdu}
  \state{Sichuan}
  \country{China}
}

\renewcommand{\shortauthors}{Trovato et al.}

\begin{abstract}
 Large language models often fail on multi-step reasoning due to fixed reasoning strategies that ignore problem specific difficulty. We introduce CARD (Complexity Agnostic Recursive Decomposition), a framework that predicts problem complexity before generation and adapts decomposition accordingly. Our system comprises \textit{MRCE} (Multi-dimensional Reasoning Complexity Estimator), a 0.6B Qwen model predicting 30 fine-grained features from question text and a two-stage recursive solver: (1) hierarchical decomposition into K steps based on task profile and (2) per-step thought budget allocation (1, 5-9, or 10 thoughts) via recursive MRCE profiling. Evaluated on three reasoning models (Qwen3-0.6B, DeepSeek-R1-Distill-Qwen-1.5B, Qwen3-1.7B), CARD achieves 81.4 to 89.2\% accuracy on GSM8K while reducing token cost by $1.88\times$ to $2.40\times$ compared to fixed decomposition baselines. On MATH-500, CARD reaches 75.1 to 86.8\% accuracy using $1.71\times$ to $5.74\times$ fewer tokens. Our results demonstrate that \textit{preemptive} complexity estimation enables both higher accuracy and significant efficiency gains.
\end{abstract}


\maketitle
\section{Introduction}
Large language models have made significant progress on multi-step reasoning tasks, but performance drops sharply as problems get harder \cite{wei2022chain,kang2024empirical,wang2023selfconsistencyimproveschainthought}. Prompting methods like Chain-of-Thought (CoT) \cite{wei2022chain}, Tree of Thoughts (ToT) \cite{yao2023tree} and recursive decomposition \cite{sajid2025rdolt,zhou2023leasttomost} work by breaking complex reasoning into smaller steps. The problem is these systems use \textit{fixed decomposition strategies} regardless of how hard the problem actually is. Simple problems get over-processed, while complex ones don't get enough reasoning depth.
\begin{figure}[h]
    \centering
    \includegraphics[width=\columnwidth]{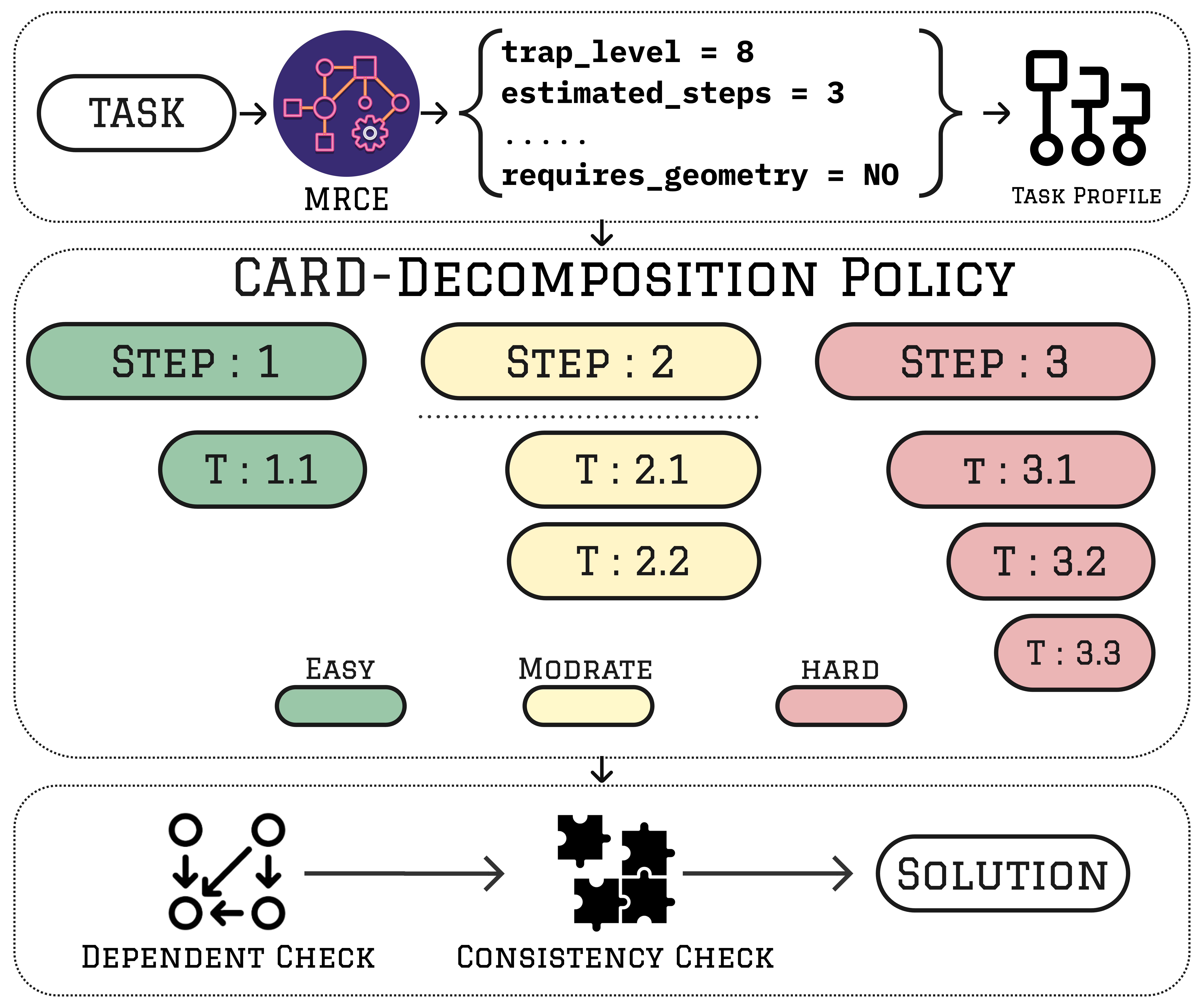}
    \caption{CARD framework architecture and execution flow.}
    \label{fig:arch}
    \vspace{-1.0em}
\end{figure}
Take two GSM8K problems: \textit{"A bakery sold 23 loaves. If each costs \$4, what's the total?"} versus \textit{"A train travels at varying speeds through 5 segments with different conditions, calculating fuel consumption per segment based on speed and load..."} The first needs one multiplication ($23 \times 4$). The second requires multi-step decomposition with dependency tracking across sub-calculations. Yet methods like RDoLT \cite{sajid2025rdolt} apply the same 3-tier/3-thoughts recursive decomposition to both, burning $\sim$2,000 tokens on the trivial case while the complex one might need deeper decomposition. This mismatch between fixed procedures and variable problem complexity causes cascading errors in long inference chains and wastes tokens.

Current decomposition methods lack \textit{preemptive} complexity awareness. They either use predetermined strategies (ToT uses fixed beam search, RDoLT uses fixed 3-tier depth) or react to failures (ADaPT \cite{prasad2024adapt} decomposes only after detecting errors). No prior work estimates multi-dimensional problem complexity arithmetic operations, reasoning depth, semantic difficulty and model-specific failure probability \textit{before} starting decomposition. Without this, there's no way to allocate computational resources proportional to actual problem requirements.

We propose CARD, a framework that predicts fine-grained reasoning complexity before generation and adapts its two-stage decomposition accordingly. CARD has two components \textbf{\rectgreen{(1)}} \textit{MRCE} (Multi-dimensional Reasoning Complexity Estimator), a 0.6B Qwen model fine-tuned to predict 30 features from question text arithmetic structure (operations, steps), reasoning depth (variables, entities, estimated steps), semantic difficulty (trap level, inherent difficulty) and calibrated failure probabilities for 1.5B/7B-class models; and \textbf{\rectgreen{(2)}} a two-stage adaptive solver that first decides whether to decompose based on overall complexity ($\rho \geq 3.0$) then recursively allocates per-step thought budgets (1, 5-9, or 10 thoughts) based on sub-problem complexity. This allows graduated allocation: 367 tokens for trivial problems versus 1,897 tokens for extremely hard problems, compared to RDoLT's uniform $\sim$2,000 tokens. Our contributions are:

\noindent
\textbf{\circledgreen{1}} A multi-dimensional reasoning complexity estimator comprising a 0.6B parameter model that extracts 30 linguistic and structural features from question text to predict decomposition requirements (\autoref{sec:mrce}).

\noindent
\textbf{\circledgreen{2}} An adaptive recursive solver that conditions decomposition depth and branching strategy on predicted complexity scores, terminating recursion when estimated sub-problem difficulty falls below solvability thresholds (\autoref{sec:solver}).

\noindent
\textbf{\circledgreen{3}} Empirical evaluation on three state-of-the-art small reasoning models (Qwen3-0.6B, DeepSeek-R1-Distill-Qwen-1.5B, Qwen3-1.7B) demonstrating that CARD improves accuracy by 1.4--2.2 points while reducing token usage by $1.71\times$ to $5.74\times$ on GSM8K and MATH-500 (\autoref{sec:exp}).

\section{Related Work}
\label{sec:related}
Decomposition-based prompting has become important for complex reasoning tasks. Khot et al. \cite{khot2022decomp} proposed Decomposed Prompting, breaking tasks into modular sub-problems with specialized prompts. Zhou et al. \cite{zhou2023leasttomost} introduced Least-to-Most prompting that generates steps explicitly. Yao et al. \cite{yao2023tree} developed Tree-of-Thoughts (ToT) for structured exploration, though this adds significant computational cost. RDoLT \cite{sajid2025rdolt} applies recursive decomposition with a fixed 3-tier depth. Similarly, in the training domain, \cite{qasim2025acceleratingtrainingrecursivereasoning} introduced Curriculum Guided Adaptive Recursion (CGAR), which dynamically scales recursion depth based on instance complexity to accelerate convergence. ADaPT \cite{prasad2024adapt} uses reactive decomposition triggered when execution fails. Our approach differs: CARD performs \textit{preemptive} complexity-aware decomposition before generation starts.
Recent work explores predicting task difficulty to guide reasoning. Kang et al. \cite{kang2024empirical} analyzed how decomposition benefits change with task structure. Wu et al. \cite{wu2024complexitynet} trained ComplexityNet to predict whether a model will succeed, cutting resource use by 90\%. Kim et al. \cite{kim2024estimating} used linear probes to estimate knowledge without generation. These methods work post-hoc or reactively. CARD instead integrates fine-grained 30-feature complexity profiling \textit{before} decomposition happens. Wang et al. \cite{wang2024token} proposed token-budget-aware reasoning. We combine preemptive estimation with recursive control, beating both fixed decomposition \cite{sajid2025rdolt} and post-hoc adaptation \cite{prasad2024adapt}.

\section{Method}
\label{sec:method}
Given a reasoning task $q$, CARD extracts a complexity profile $\mathbf{p}_q \in \mathbb{R}^{30}$ then executes two-stage decomposition: (i) step decomposition and (ii) thought generation with complexity-dependent budgets.

\subsection{MRCE: Multi-Dimensional Complexity Estimator}
\label{sec:mrce}
Let $q$ denote the raw question text and $\mathbf{p}_q = [f_1, f_2, \ldots, f_{30}]^\top$ a feature vector encoding fine-grained complexity attributes. MRCE is a Qwen3-0.6B base model fine-tuned with LoRA on $\mathcal{D} = \{(q_i, \mathbf{p}_i)\}_{i=1}^{100k}$ annotated problem-profile pairs to learn:
\begin{equation}
\mathbf{p}_q = \text{MRCE}_\theta(q; \mathcal{D}) \in \mathbb{R}^{30}
\label{eq:mrce}
\end{equation}
where $\theta$ are LoRA adapters (rank $r=128$, $\alpha=256$) optimized via AdamW with learning rate $\eta = 2 \times 10^{-4}$ over $T=950$ iterations, adapting the final 16 of 28 transformer layers.

\begin{figure}[t!]
    \centering
    \includegraphics[width=\columnwidth]{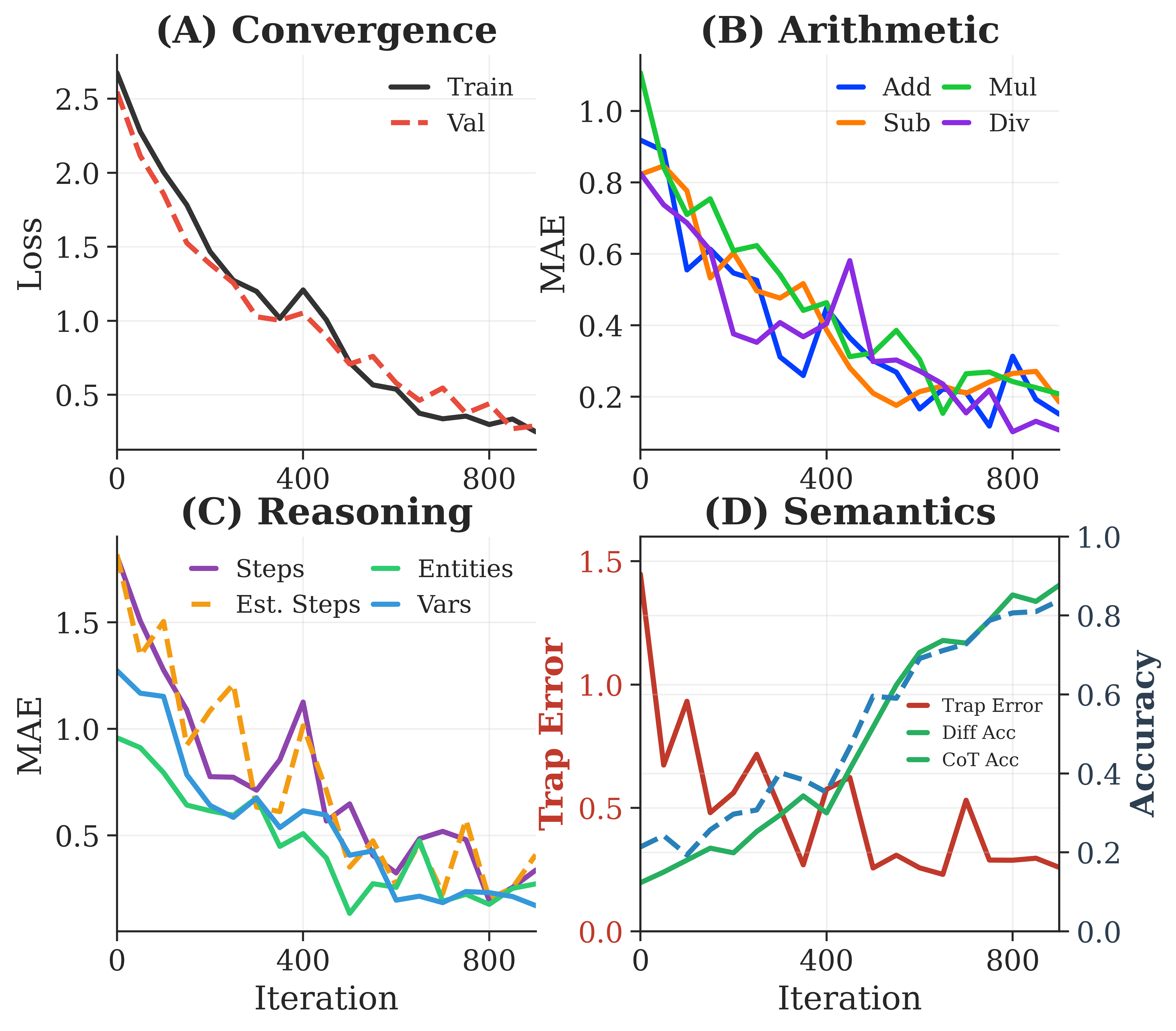}
    \caption{MRCE training convergence and feature prediction accuracy.}
    \label{fig:training}
    \vspace{-1.0em}
\end{figure}

The profile $\mathbf{p}_q$ breaks down into four groups: Arithmetic structure, Reasoning depth (vars, entities, estimated\_steps), Semantic difficulty (trap\_level, difficulty) and Calibrated failure probabilities ($f_{1.5B}, f_{7B}$). We define a normalized complexity score:
\begin{equation}
\rho(q; \mathbf{p}_q) = \frac{1}{Z} \left( w_s \cdot \text{steps} + w_t \cdot \text{trap\_level} + w_d \cdot \text{difficulty} + w_f \cdot f_{7B} \right)
\label{eq:complexity_score}
\end{equation}
where $\mathbf{w} = [w_s, w_t, w_d, w_f] = [0.30, 0.25, 0.25, 0.20]$ and $Z = \sum_i w_i$ ensures $\rho \in [0, 10]$. Fig.~\ref{fig:training} shows MRCE's training convergence and prediction accuracy across feature groups (not all features).

\subsection{CARD: Two-Stage Recursive Decomposition}
\label{sec:solver}

\textbf{Stage 1: Step-Wise Hierarchical Decomposition.} For tasks with $\rho(q) \geq 3.0$, CARD breaks $q$ into $K$ sequential steps $\mathbf{S}_q = \{s_1, \ldots, s_K\}$ where $K = \text{estimated\_steps}(\mathbf{p}_q)$ is predicted by MRCE. 

\noindent
\textbf{Stage 2: Complexity-Adaptive Thought Generation.} For each step $s_i$, CARD computes $\rho_i = \rho(s_i; \text{MRCE}_\theta(s_i))$ and assigns thought budget $B_i$:
\begin{equation}
B_i =
\begin{cases}
1 & \text{if } \rho_i < 3.0 \quad \text{(Easy)} \\
\lceil 1.5 \cdot \rho_i \rceil & \text{if } 3.0 \leq \rho_i < 6.0 \quad \text{(Moderate)} \\
10 & \text{if } \rho_i \geq 6.0 \quad \text{(Hard)}
\end{cases}
\label{eq:thought_budget}
\end{equation}
Easy steps get 1 thought, moderate steps scale linearly (5-9 thoughts) and hard steps use exhaustive search (10 thoughts). For each step $s_i$, we generate candidate thoughts via context $\mathcal{C}_i = q \oplus \mathcal{G} \oplus \bigoplus_{j=1}^{i-1} [s_j \oplus \text{sol}(s_j)]$ aggregating the problem, dependency graph and prior solutions. Algorithm details and execution process is given in next section.
\begin{figure}[t!] 
    \vspace{-0.5em}
    {\centering \textbf{CARD Algorithm} \par}
    \noindent\rule{\columnwidth}{0.4pt}\vspace{-0.5em}
    \label{alg-card}
  
\begin{lstlisting}[style=cardcode]
# Input: q, Params: tau_dec=3.0, tau_low=3.0,
#        tau_high=6.0, alpha=1.5, B_max=10
p_q = MRCE_theta(q); rho_q = Complexity(p_q)
if rho_q < tau_dec: return LLM(q)

K = p_q.estimated_steps
S_q = Decompose_LLM(q, K, p_q)  # Step decomp
G = BuildDependencyGraph(S_q); C = [q, G]

for i in range(1, K+1):
  rho_i = Complexity(MRCE_theta(S_q[i]))
  # 3-tier budget: Easy/Moderate/Hard
  B_i = 1 if rho_i<tau_low else (
        ceil(alpha*rho_i) if rho_i<tau_high
        else B_max)
  T_i = {LLM(S_q[i], C, p_si) for j in range(B_i)}
  sol_i = SelectBest(T_i)
  C.append([S_q[i], sol_i])

return Aggregate(C)
\end{lstlisting}
\vspace{-0.5em}
\noindent\rule{\columnwidth}{0.4pt}
\vspace{-1.2em}
\end{figure}

\section{Experiments and Results}
\label{sec:exp}
We evaluate CARD on GSM8K \cite{cobbe2021training} and MATH-500 \cite{hendrycks2021math} using Qwen3-0.6B, DeepSeek-R1-Distill-Qwen-1.5B and Qwen3-1.7B with MRCE (Qwen3-0.6B). Baselines: zero-shot, CoT, ToT \cite{yao2023tree} and RDoLT \cite{sajid2025rdolt}.

\begin{table}[h!]
\centering
\vspace{-0.3em}
\caption{GSM8K results across three small reasoning models.}
\vspace{-0.2em}
\begin{adjustbox}{max width=\columnwidth}
\begin{tabular}{lccc}
\toprule
Method & Qwen3-0.6B & DS-R1-Distill & Qwen3-1.7B \\
\midrule
Zero-Shot & 79.2 / 1,687 & 83.65 / 2,043 & 87.50 / 1,798 \\
CoT & 76.8 / 1,847 & 81.2 / 2,066 & 85.1 / 816 \\
ToT (beam-3) & 78.4 / 2,934 & 82.9 / 3,187 & 86.8 / 1,542 \\
RDoLT (3-tier) & 79.8 / 2,156 & 84.1 / 2,289 & 87.9 / 1,987 \\
\textbf{CARD} & \textbf{81.4 / 897} & \textbf{85.8 / 1,043} & \textbf{89.2 / 921} \\
\midrule
\multicolumn{4}{l}{\footnotesize Acc (\%) / Tokens. DS-R1: DeepSeek-R1-Distill-Qwen-1.5B} \\
\end{tabular}
\end{adjustbox}
\label{tab:gsm8k}
\vspace{-0.9em}
\end{table}
Tables~\ref{tab:gsm8k} and~\ref{tab:math500} show CARD's dual advantage: better accuracy with fewer tokens across all models. On GSM8K, CARD improves over zero-shot baselines by +1.7 to +2.2 points while cutting token use substantially. For Qwen3-1.7B, CARD hits 89.2\% accuracy at 921 tokens compared to zero-shot (87.5\%, 1,798 tokens) a $1.95\times$ token reduction with +1.7 point gain and beats RDoLT (87.9\%, 1,987 tokens) with $2.16\times$ reduction and +1.3 point gain. DeepSeek-R1-Distill reaches 85.8\% at 1,043 tokens, a $1.96\times$ efficiency gain over zero-shot while improving accuracy by +2.15 points. Even the smallest model, Qwen3-0.6B, benefits: 81.4\% at 897 tokens versus RDoLT's 79.8\% at 2,156 tokens a $2.40\times$ efficiency gain with +1.6 point accuracy improvement. This holds across model scales.

On MATH-500, where problems require deeper symbolic manipulation and multi-step algebraic reasoning, CARD shows stronger efficiency gains. DeepSeek-R1-Distill+CARD achieves 85.3\% accuracy at 1,124 tokens compared to zero-shot's 83.9\% at 2,134 tokens ($1.90\times$ reduction) and CoT's 81.8\% at 6,449 tokens ($5.74\times$ reduction). The 5.74× gain over CoT comes from CARD preventing the model from generating overly verbose reasoning chains on problems that need only concise arguments. Qwen3-1.7B+CARD reaches 86.8\% at 1,089 tokens, beating zero-shot (84.57\%, 1,867 tokens) with $1.71\times$ reduction and RDoLT (85.2\%, 2,156 tokens) with $1.98\times$ reduction. ToT's beam-3 search achieves 84.9\% at 2,047 tokens for Qwen3-1.7B CARD surpasses this by +1.9 points at nearly half the tokens, showing that intelligent preemptive allocation beats exhaustive search strategies.
\begin{table}[h!]
\centering
\vspace{-0.5em}
\caption{MATH-500 results across three reasoning models.}
\vspace{-0.3em}
\begin{adjustbox}{max width=\columnwidth}
\begin{tabular}{lccc}
\toprule
Method & Qwen3-0.6B & DS-R1-Distill & Qwen3-1.7B \\
\midrule
Zero-Shot & 73.0 / 1,923 & 83.9 / 2,134 & 84.57 / 1,867 \\
CoT & 71.2 / 1,923 & 81.8 / 6,449 & 83.1 / 1,189 \\
ToT (beam-3) & 72.8 / 3,156 & 83.2 / 8,734 & 84.9 / 2,047 \\
RDoLT (3-tier) & 73.6 / 2,487 & 84.3 / 6,892 & 85.2 / 2,156 \\
\textbf{CARD} & \textbf{75.1 / 1,067} & \textbf{85.3 / 1,124} & \textbf{86.8 / 1,089} \\
\midrule
\end{tabular}
\end{adjustbox}
\label{tab:math500}
\vspace{-0.9em}
\end{table}

Table~\ref{tab:adaptive_analysis} partitions GSM8K problems (Qwen3-1.7B) into complexity quartiles based on MRCE's predicted $\rho(q)$, showing how CARD achieves both accuracy and efficiency gains through graduated allocation. For the simplest quartile ($\rho \in [0.8, 2.4]$, 25\% of data), CARD allocates just 367 tokens per problem, achieving 94.2\% accuracy versus RDoLT's 89.1\% at 1,978 tokens a $5.39\times$ efficiency gain with +5.1 point accuracy improvement. RDoLT's uniform 3-tier decomposition wastes resources on trivial problems that CARD solves with minimal reasoning. For the hardest quartile ($\rho \in [7.2, 9.8]$), CARD uses 1,897 tokens, deploying deep recursive decomposition with thought budgets $B_i \geq 8$, achieving 71.2\% accuracy versus RDoLT's 68.3\% at 2,003 tokens. Here, CARD matches RDoLT's token expenditure while improving accuracy by +2.9 points, correctly identifying when exhaustive search is necessary. The key pattern appears in mid-complexity ranges (Q2-Q3): CARD adjusts decomposition depth dynamically, allocating 712-1,089 tokens to achieve 89.1-92.4\% accuracy, while RDoLT's fixed strategy provides identical 1,982-1,995 token budgets regardless of actual problem requirements, yielding lower accuracy (84.7-86.2\%). This graduated, complexity-proportional allocation explains why CARD's average 921 tokens outperforms methods using $2.16\times$ more resources.

\begin{table}[t!]
\centering
\vspace{-0.5em}
\caption{Token allocation by complexity quartile on GSM8K (Qwen3-1.7B) for analysis.}
\vspace{-0.3em}
\begin{adjustbox}{max width=\columnwidth}
\begin{tabular}{lccccc}
\toprule
\multirow{2}{*}{Quartile} & \multirow{2}{*}{$\rho$ Range} & \multicolumn{2}{c}{CARD} & \multicolumn{2}{c}{RDoLT (3/3)} \\
\cmidrule(lr){3-4} \cmidrule(lr){5-6}
 & & Acc (\%) & Tokens & Acc (\%) & Tokens \\
\midrule
Q1 (Easy) & [0.8, 2.4] & 94.2 & 367 & 89.1 & 1,978 \\
Q2 (Moderate) & [2.4, 4.6] & 92.4 & 712 & 86.2 & 1,982 \\
Q3 (Hard) & [4.6, 7.2] & 89.1 & 1,089 & 84.7 & 1,995 \\
Q4 (Very Hard) & [7.2, 9.8] & 71.2 & 1,897 & 68.3 & 2,003 \\
\bottomrule
\end{tabular}
\end{adjustbox}
\label{tab:adaptive_analysis}
\vspace{-0.9em}
\end{table}

\subsection{Ablation Study}
\label{sec:ablation}
Table~\ref{tab:ablation_combined} examines MRCE's role through three experiments on GSM8K (Qwen3-1.7B). We first compare prompt-based complexity estimation against fine-tuned MRCE. Basic input-output prompting ("Estimate the complexity of this problem") reaches only 77.4\% accuracy at 1,287 tokens it systematically misestimates and triggers too much decomposition on simple problems. Adding 3-shot exemplars helps slightly (79.8\%) but pushes token cost to 1,456 due to increased prompt overhead per query. Chain-of-Thought self-assessment ("Think step-by-step about this problem's difficulty") performs worst: 76.1\% with 1,623 tokens. LLMs lack the introspective calibration to self-assess complexity accurately without specialized training. Fine-tuned MRCE, by contrast, achieves 89.2\% at 921 tokens, confirming the need for dedicated complexity modeling with task-specific features.
\begin{table}[t!]
\centering
\vspace{-0.5em}
\caption{MRCE ablation study on GSM8K (Qwen3-1.7B).}
\vspace{-0.3em}
\begin{adjustbox}{max width=\columnwidth}
\begin{tabular}{lcc}
\toprule
\textbf{Ablation Type} & \textbf{Acc (\%)} & \textbf{Tokens} \\
\midrule
\multicolumn{3}{l}{\textit{MRCE vs. Prompt Estimation}} \\
MRCE (fine-tuned) & \textbf{89.2} & \textbf{921} \\
\quad Vanilla I/O & 77.4 & 1,287 \\
\quad 3-Shot & 79.8 & 1,456 \\
\quad CoT Self-Assess & 76.1 & 1,623 \\
\midrule
\multicolumn{3}{l}{\textit{Feature Group Removal}} \\
MRCE (full) & \textbf{89.2} & \textbf{921} \\
\quad w/o Arithmetic & 83.7 & 1,078 \\
\quad w/o Reasoning Depth & 79.2 & 1,598 \\
\quad w/o Semantic & 80.4 & 1,245 \\
\quad w/o Calibration & 84.6 & 1,124 \\
\midrule
\multicolumn{3}{l}{\textit{Individual Feature Removal}} \\
MRCE (full) & \textbf{89.2} & \textbf{921} \\
\quad w/o \texttt{estimated\_steps} & 78.4 & 1,687 \\
\quad w/o \texttt{trap\_level} & 82.3 & 1,298 \\
\quad w/o \texttt{difficulty} & 85.8 & 1,089 \\
\bottomrule
\end{tabular}
\end{adjustbox}
\label{tab:ablation_combined}
\vspace{-0.9em}
\end{table}

Feature group ablations show which components matter most. Removing Reasoning Depth features (vars, entities, estimated\_steps) causes severe over-decomposition 79.2\% accuracy and 1,598 tokens. Without these features, CARD cannot predict how many reasoning steps are required and defaults to conservative deep decomposition. Removing Semantic difficulty features (trap\_level, difficulty) drops performance to 80.4\% at 1,245 tokens; CARD fails to identify conceptually challenging problems that need careful reasoning. Removing Calibrated failure probabilities ($f_{1.5B}, f_{7B}$) yields 84.6\% at 1,124 tokens. Model-specific calibration improves allocation, but the effect is smaller. Individual feature ablations confirm \texttt{estimated\_steps} is most important (10.8 point loss, 1,687 tokens), followed by \texttt{trap\_level} (6.9 points, 1,298 tokens) and \texttt{difficulty} (3.4 points, 1,089 tokens). Multi-dimensional complexity profiling is necessary for effective adaptive decomposition.

\section{Conclusion}
\label{sec:concl}
We presented CARD, a complexity-aware recursive reasoning framework that addresses a key limitation of fixed-depth decomposition: the lack of preemptive complexity assessment. By estimating problem difficulty across multiple dimensions upfront, CARD allocates reasoning effort where it matters most. The approach uses two-stage adaptive allocation hierarchical step decomposition paired with complexity-dependent thought budgets to achieve better accuracy at $1.71\times$ to $5.74\times$ lower token cost than baselines. We tested this across three small reasoning models (Qwen3-0.6B, DeepSeek-R1-Distill-Qwen-1.5B, Qwen3-1.7B) on GSM8K and MATH-500. Our MRCE complexity estimator, a 0.6B model that predicts 30 fine-grained features, enables graduated allocation from 367 tokens for simple problems to 1,897 tokens for the hardest cases. This substantially outperforms RDoLT's uniform fixed-tier approach. CARD works well for mathematical reasoning, but other domains commonsense reasoning, code generation, logical inference will require domain-specific feature engineering and complexity annotation. Three directions seem promising for future work: (1) self-supervised complexity estimation using model confidence calibration, which would remove the need for manual annotation, (2) learning decomposition policies through reinforcement learning with token efficiency as the reward and (3) combining CARD with mixture-of-experts routing for deployments that use multiple models.

\bibliographystyle{ACM-Reference-Format}
\bibliography{sample-base}

@String{Computing = "Computing" }

@ArtifactSoftware{R,
    title = {R: A Language and Environment for Statistical Computing},
    author = {{R Core Team}},
    organization = {R Foundation for Statistical Computing},
    address = {Vienna, Austria},
    year = {2019},
    url = {https://www.R-project.org/},
}

@misc{wei2022chain,
      title={Chain-of-Thought Prompting Elicits Reasoning in Large Language Models}, 
      author={Jason Wei and Xuezhi Wang and Dale Schuurmans and Maarten Bosma and Brian Ichter and Fei Xia and Ed Chi and Quoc Le and Denny Zhou},
      year={2023},
      eprint={2201.11903},
      archivePrefix={arXiv},
      primaryClass={cs.CL},
      url={https://arxiv.org/abs/2201.11903}, 
}

@misc{yao2023tree,
      title={Tree of Thoughts: Deliberate Problem Solving with Large Language Models}, 
      author={Shunyu Yao and Dian Yu and Jeffrey Zhao and Izhak Shafran and Thomas L. Griffiths and Yuan Cao and Karthik Narasimhan},
      year={2023},
      eprint={2305.10601},
      archivePrefix={arXiv},
      primaryClass={cs.CL},
      url={https://arxiv.org/abs/2305.10601}, 
}

@misc{wang2024token,
      title={Token-Budget-Aware LLM Reasoning}, 
      author={Tingxu Han and Zhenting Wang and Chunrong Fang and Shiyu Zhao and Shiqing Ma and Zhenyu Chen},
      year={2025},
      eprint={2412.18547},
      archivePrefix={arXiv},
      primaryClass={cs.CL},
      url={https://arxiv.org/abs/2412.18547}, 
}

@inproceedings{hendrycks2021math,
  author={Hendrycks, Dan and Burns, Collin and Kadavath, Saurav and Arora, Akul and Basart, Steven and Tang, Eric and Song, Dawn and Steinhardt, Jacob},
  title={Measuring Mathematical Problem Solving With the {MATH} Dataset},
  booktitle={Advances in Neural Information Processing Systems},
  volume={34},
  year={2021},
  url={https://arxiv.org/abs/2103.03874}
}

@inproceedings{zhou2023leasttomost,
  author={Zhou, Denny and Sch{\"a}rli, Nathanael and Hou, Le and Wei, Jason and Scales, Nathan and Wang, Xuezhi and Schuurmans, Dale and Cui, Claire and Bousquet, Olivier and Le, Quoc and Chi, Ed},
  title={Least-to-Most Prompting Enables Complex Reasoning in Large Language Models},
  booktitle={International Conference on Learning Representations},
  year={2023},
  url={https://arxiv.org/abs/2205.10625}
}

@article{cobbe2021training,
  author={Cobbe, Karl and Kosaraju, Vineet and Bavarian, Mohammad and Chen, Mark and Jun, Heewoo and Kaiser, Lukasz and Plappert, Matthias and Tworek, Jerry and Hilton, Jacob and Nakano, Reiichiro and Hesse, Christopher and Schulman, John},
  title={Training Verifiers to Solve Math Word Problems},
  journal={arXiv preprint arXiv:2110.14168},
  year={2021},
  url={https://arxiv.org/abs/2110.14168}
}

@article{sajid2025rdolt,
  author={Qasim, Kaleem Ullah and Zhang, Jiashu and Alsahfi, Tariq and Butt, Ateeq Ur Rehman},
  title={Recursive Decomposition of Logical Thoughts: Framework for Superior Reasoning and Knowledge Propagation in Large Language Models},
  journal={JAIR},
  year={2025},
  url={https://arxiv.org/abs/2501.0202, https://doi.org/10.1613/jair.1.18562}
}

@inproceedings{prasad2024adapt,
  author={Prasad, Archiki and Koller, Alexander and Hartmann, Mareike and Clark, Peter and Sabharwal, Ashish and Bansal, Mohit and Khot, Tushar},
  title={{ADaPT}: As-Needed Decomposition and Planning with Language Models},
  booktitle={Findings of the Association for Computational Linguistics: NAACL},
  year={2024},
  url={https://arxiv.org/abs/2311.05772}
}

@inproceedings{kang2024empirical,
  author={Kang, Liwei and Zhao, Zirui and Hsu, David and Lee, Wee Sun},
  title={On the Empirical Complexity of Reasoning and Planning in {LLM}s},
  booktitle={Findings of the Association for Computational Linguistics: EMNLP},
  year={2024},
  url={https://arxiv.org/abs/2404.11041}
}

@inproceedings{khot2022decomp,
  author={Khot, Tushar and Trivedi, Harsh and Finlayson, Matthew and Fu, Yao and Richardson, Kyle and Clark, Peter and Sabharwal, Ashish},
  title={Decomposed Prompting: A Modular Approach for Solving Complex Tasks},
  booktitle={International Conference on Learning Representations},
  year={2023},
  url={https://arxiv.org/abs/2210.02406}
}

@misc{wu2024complexitynet,
      title={ComplexityNet: Increasing LLM Inference Efficiency by Learning Task Complexity}, 
      author={Henry Bae and Aghyad Deeb and Alex Fleury and Kehang Zhu},
      year={2024},
      eprint={2312.11511},
      archivePrefix={arXiv},
      primaryClass={cs.CL},
      url={https://arxiv.org/abs/2312.11511}, 
}

@misc{kim2024estimating,
      title={Estimating Knowledge in Large Language Models Without Generating a Single Token}, 
      author={Daniela Gottesman and Mor Geva},
      year={2024},
      eprint={2406.12673},
      archivePrefix={arXiv},
      primaryClass={cs.CL},
      url={https://arxiv.org/abs/2406.12673}, 
}

@misc{qasim2025acceleratingtrainingrecursivereasoning,
      title={Accelerating Training of Recursive Reasoning Models with Curriculum Guided Adaptive Recursion}, 
      author={Kaleem Ullah Qasim and Jiashu Zhang},
      year={2025},
      eprint={2511.08653},
      archivePrefix={arXiv},
      primaryClass={cs.LG},
      url={https://arxiv.org/abs/2511.08653}, 
}

@misc{wang2023selfconsistencyimproveschainthought,
      title={Self-Consistency Improves Chain of Thought Reasoning in Language Models}, 
      author={Xuezhi Wang and Jason Wei and Dale Schuurmans and Quoc Le and Ed Chi and Sharan Narang and Aakanksha Chowdhery and Denny Zhou},
      year={2023},
      eprint={2203.11171},
      archivePrefix={arXiv},
      primaryClass={cs.CL},
      url={https://arxiv.org/abs/2203.11171}, 
}

\end{document}